\documentclass[sigconf]{acmart}

\AtBeginDocument{%
  }

\usepackage{enumitem}
\usepackage{multirow}
\usepackage{kotex}
\usepackage{mathtools}
\usepackage{booktabs}
\usepackage{makecell}
\usepackage{kotex}
\usepackage{pifont}
\usepackage{tabularx}
\usepackage{balance}
\usepackage{xcolor}

\copyrightyear{2026}
\acmYear{2026}
\setcopyright{cc}
\setcctype{by}
\acmConference[CIKM '26]{Proceedings of the 35th ACM International Conference on Information and Knowledge Management}{November 07--11, 2026}{Rome, Italy}
\acmBooktitle{Proceedings of the 35th ACM International Conference on Information and Knowledge Management (CIKM '26), November 07--11, 2026, Rome, Italy}
\acmDOI{10.1145/3799682.3839914}
\acmISBN{979-8-4007-2539-5/2026/11}

\makeatletter
\gdef\@copyrightpermission{
\begin{minipage}{0.3\columnwidth}
\href{https://creativecommons.org/licenses/by/4.0/}{\includegraphics[width=0.90\textwidth]{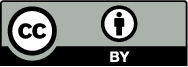}}
\end{minipage}\hfill
\begin{minipage}{0.7\columnwidth}
\href{https://creativecommons.org/licenses/by/4.0/}{This work is licensed under a Creative Commons Attribution International 4.0 License.}
\end{minipage}
\vspace{5pt}
}
\makeatother

\begin{document}

\title{A Behavioral Trait Leaks into Preferences: Diagnosing Trait Interference in LLM User Simulators}

\author{Chaehyun Kim}
\email{ch.kim@kaist.ac.kr}
\affiliation{
\institution{KAIST}
\city{Daejeon}
\country{Republic of Korea}
}

\author{Sein Kim}
\email{rlatpdlsgns@kaist.ac.kr}
\affiliation{
\institution{KAIST}
\city{Daejeon}
\country{Republic of Korea}
}

\author{Hongseok Kang}
\email{ghdtjr0311@kaist.ac.kr}
\affiliation{
\institution{KAIST}
\city{Daejeon}
\country{Republic of Korea}
}

\author{Chanyoung Park}
\authornote{Corresponding author.}
\email{cy.park@kaist.ac.kr}
\affiliation{
\institution{KAIST}
\city{Daejeon}
\country{Republic of Korea}
}

\renewcommand{\shortauthors}{Chaehyun Kim, Sein Kim, Hongseok Kang, and Chanyoung Park}



\begin{abstract}
    LLM-based user simulators aim to bridge the offline–online gap in recommender evaluation by emulating users through injected traits, where preference attributes determine what a user engages with and a behavioral activity trait governs how long they browse. However, we show this intended trait independence collapses during simulation, causing two failures: (i) \textbf{Trait Interference}, where amplified activity distorts preference boundaries and forces interactions with mismatched items to sustain browsing, and (ii) \textbf{Evaluation Invalidity}, where satisfaction scores inflate with activity-driven page counts despite taste mismatches, biasing evaluation toward trait distributions rather than recommender performance. To resolve this, we propose PQA, a page-level quality anchoring method that guides simulators using a personalized anchor reflecting each user's intrinsic preference standard. By assessing whether a page meets this standard before further browsing, PQA enables proactive exits from low-quality pages, letting the activity trait retain its intended role of modulating browsing depth within preference-conforming pages. Experiments show PQA mitigates trait interference and improves the reliability of LLM-based simulator evaluation under activity shifts.
    Our code is available at \url{https://github.com/chaehyun1/PQA}.
\end{abstract}




\maketitle

\section{Introduction}
\label{sec: introduction}

LLM-based user simulators have recently been proposed to replicate realistic user behaviors by leveraging the reasoning ability of LLMs~\cite{wang2025user,zhang2024generative,bougie2025simuser,huang2025recommender,wang2024recmind}.
To construct more realistic virtual users, existing simulators define user traits and inject them into the input prompts, where the simulator browses a recommendation list presented page by page and decides when to stop~\cite{zhang2024generative, bougie2025simuser}. These traits fall into preference attributes (e.g., rating conformity and category diversity), which define \textit{what} a user likes, and a behavioral trait such as user activity, which is designed to determine only \textit{how long} a user keeps browsing the recommendation pages~\cite{zhang2024generative, 10.1145/2187980.2188230, bougie2025simuser}.

Despite this design, whether the activity trait preserves its role during simulation has been largely overlooked. Through a series of analyses, we find that current simulators fail to keep the activity trait within its intended role, leading to two critical issues:
\begin{enumerate}[leftmargin=0.38cm]
    \item \textbf{Trait Interference:} 
    As the activity trait is amplified, the simulator increasingly selects items semantically distant from the user's persona (Section~\ref{sec: persona-item embedding similarity}) and deviates from the user's intrinsic category preferences (Section~\ref{sec: category overlap}). This contradicts real-world data, where user activity is negatively correlated with average ratings (Section~\ref{sec: Correlation between Activity and Rating}): more active users tend to give lower ratings on average, indicating stricter rather than permissive behavior. By contrast, the simulator indiscriminately interacts with mismatched items strictly to fulfill the behavioral constraint of extended browsing.
    {\item \textbf{Evaluation Invalidity:} 
    Unlike preference attributes (i.e., conformity and diversity), whose amplification has little effect on the number of explored pages and the overall satisfaction score, activity amplification sharply increases both metrics (Section~\ref{sec: trait comparison}). This inflation persists even when recommendation quality is degraded, as high-activity simulators continue browsing low-quality pages despite preference mismatches. As a result, the satisfaction score becomes confounded by the injected activity level rather than reflecting the actual performance of the recommender system.}
\end{enumerate}

We identify the root cause of the aforementioned trait interference and subsequent evaluation invalidity as the absence of an explicit page-level quality anchor within the simulator, allowing activity constraints to override page-level preference alignment.
{Ideally, the activity trait should govern how extensively a user explores, without relaxing the user's preference boundary. This interpretation is consistent with prior simulator designs. Specifically, Agent4Rec~\cite{zhang2024generative} encodes personalized preferences from historical likes and dislikes, summarizing them into unique tastes and rating patterns, while using the activity trait for exit decisions. Similarly, SimUSER~\cite{bougie2025simuser} defines the corresponding behavioral trait as the frequency and breadth of a user's interactions with recommended items. Thus, although activity may affect browsing depth and interaction frequency, it should not justify continued interaction with preference-mismatched items.}

To address this, we introduce a page-level quality anchoring method called PQA, which defines a personalized anchor $\mu_u$ that captures how densely each user's preferred categories typically appear within a page, based on their interaction history. 
By injecting $\mu_u$ into the prompt, the simulator can judge whether a recommended page meets the user's preference expectation before deciding whether to continue browsing. This enables the simulator to proactively exit low-quality pages, ensuring that the activity trait modulates browsing depth only when page quality remains acceptable.
Our main contributions are summarized as follows:

\begin{itemize}[leftmargin=0.38cm]
    \item We quantitatively demonstrate trait interference in user simulators, where changes in the activity trait distort users' preference standards and consequently undermine evaluation validity.
    \item We identify the absence of a page-level quality anchor as the root cause of trait interference and propose PQA, which uses a personalized anchor $\mu_u$ derived from each user's historical category consumption to mitigate this issue.
    \item Through extensive experiments, we validate that PQA preserves trait independence and restores the evaluation validity of the simulator under activity distribution shifts.
\end{itemize}

\section{Preliminary Analysis}
\label{sec: preliminary analysis}

\textbf{Simulation Framework.}
{A backbone recommender model (e.g., SASRec~\cite{kang2018self}, LightGCN~\cite{he2020lightgcn}, MultVAE~\cite{liang2018variational}) generates a personalized ranked item list by scoring candidate items from each user's interaction history. 
To emulate real-world browsing behavior, the list is partitioned into pages and sequentially presented to the user simulator. At each page, the simulator decides which items to interact with based on the injected persona and user traits, and then determines whether to continue to the next page or exit the session.}

\smallskip
\noindent\textbf{Evaluation Protocol.} 
{To investigate trait interference, we use Agent4Rec~\cite{zhang2024generative} and SimUSER~\cite{bougie2025simuser} with SASRec~\cite{kang2018self} as the backbone recommender on MovieLens~\cite{harper2015movielens} and Amazon CDs~\cite{mcauley2015image}. SASRec is trained with a leave-last-out strategy~\cite{kang2018self, sun2019bert4rec, 10.1145/3711896.3737035,tang2018personalized}, and the same training data are used to construct user personas and traits. For each user, we build a 20-item recommendation list with a $1:k$ ratio of positive top-ranked items to negative bottom-ranked items. Following standard interfaces~\cite{zhang2024generative, bougie2025simuser}, we present four items per page and order the list by predicted logits, so that item relevance strictly decreases with page depth. This controlled degradation allows us to test whether high activity drives the simulator to continue browsing low-quality pages beyond the user's core interests.}


\subsection{Trait Interference}
\label{sec: trait interference}

\begin{figure}[t]
  \centering
  \includegraphics[width=1\linewidth]{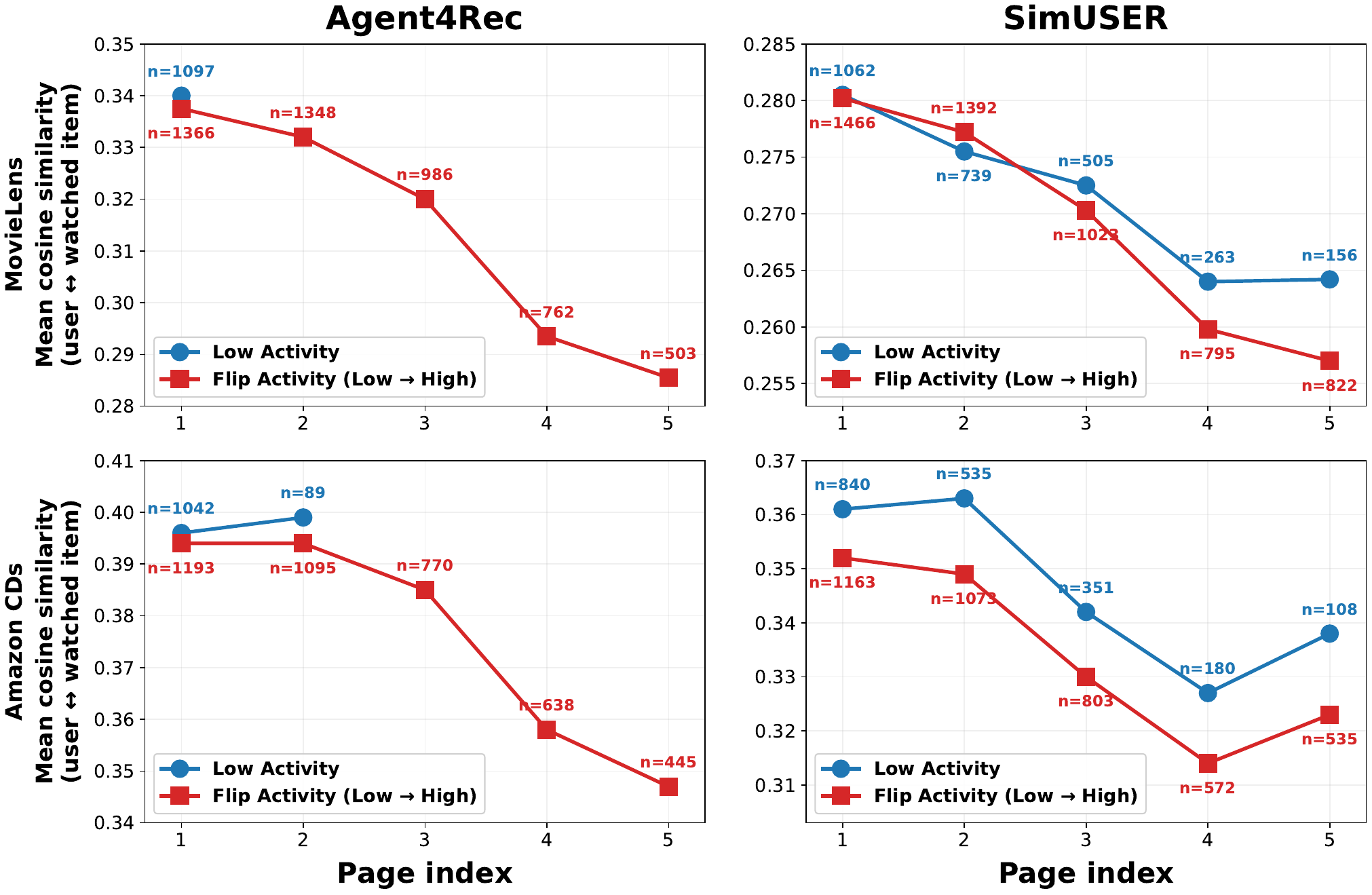}
  \caption{Mean cosine similarity between personas and watched items across page depths. $n$ denotes the total number of watched items per page (plotted for $n \ge 20$).}
    \label{fig: similarity}
\end{figure}

\subsubsection{\textbf{Persona-item embedding similarity}}\label{sec: persona-item embedding similarity}
To investigate the trait interference caused by the behavioral trait, we sample low-activity users and modify their activity level to a high-activity state while holding their preference attributes constant. For each user, the set of items presented on each page is kept identical across activity levels. Specifically, under the 1:1 setting, we evaluate the semantic alignment between the virtual user's intrinsic preferences and the simulator's decisions by computing the cosine similarity between the user persona and the textual metadata of items selected (i.e., watched) by the simulator. Both the personas and item metadata are embedded using the \texttt{all-MiniLM-L6-v2}~\cite{wang2020minilm} model.

As illustrated in Figure~\ref{fig: similarity}, amplifying the activity trait increases the overall number of interactions ($n$) across all pages and drives the simulator to explore deeper page depths. 
Crucially, the high-activity state yields lower cosine similarity than the low-activity state at the same page depth, indicating that the simulator engages with more preference-mismatched items to sustain browsing.
Indeed, as recommendation quality degrades with page depth, low-activity users show substantially fewer interactions in later pages, whereas high-activity users maintain a notably high volume of watched items ($n$) despite declining relevance. This indiscriminate acceptance provides empirical evidence of trait interference, suggesting that the LLM relaxes its preference boundaries to satisfy the behavioral constraint of extended browsing.

\subsubsection{\textbf{Category overlap}}\label{sec: category overlap}
To evaluate preference consistency during the simulation, we measure the category overlap ratio. This metric compares the simulator's page-level decisions against a personalized baseline derived from the real user's interaction history, allowing us to quantify how closely the simulation aligns with the user's intrinsic evaluation standards.

To operationalize this standard, we first compute the user's historical baseline. From the user's interaction history, we extract the top-$K$ categories (we set $K=5$) and partition the timeline into discrete sessions. 
For each session $i$, we compute the overlap as the average number of top-$K$ categories within the interacted items:
\begin{equation}
\small
    \mathrm{overlap}_i = \frac{1}{|\mathrm{session}_i|} \sum_{v \in \mathrm{session}_i} |\mathrm{categories}(v) \cap \mathrm{top\text{-}}K|
    \label{eq: overlap}
\end{equation}
where $\mathrm{categories}(v)$ denotes the set of categories associated with item $v$. 
To reflect the user's most recent persona, we apply a normalized exponential weight $w_i = \frac{\gamma^i}{\sum_j \gamma^j}$ ($\gamma = 0.9$, with $i=0$ as the most recent session). The historical baseline is then defined as:
\begin{equation}
\small
\mathrm{baseline} = \sum_i w_i \cdot \mathrm{overlap}_i 
\label{eq: baseline}
\end{equation}
We evaluate the entire page to capture recommendation quality before the simulator decides whether to continue or exit.
{During simulation, we compute page-level overlap by applying Equation~\ref{eq: overlap} with a browsed page replacing the historical session.}
The final metric is the ratio of this page-level overlap to the historical baseline ($\mathrm{page\_overlap} / \mathrm{baseline}$). A ratio approximating $1.0$ indicates that the simulator strictly adheres to the user's inherent preference boundaries. 

\begin{figure}[t]
  \centering
  \includegraphics[width=1\linewidth]{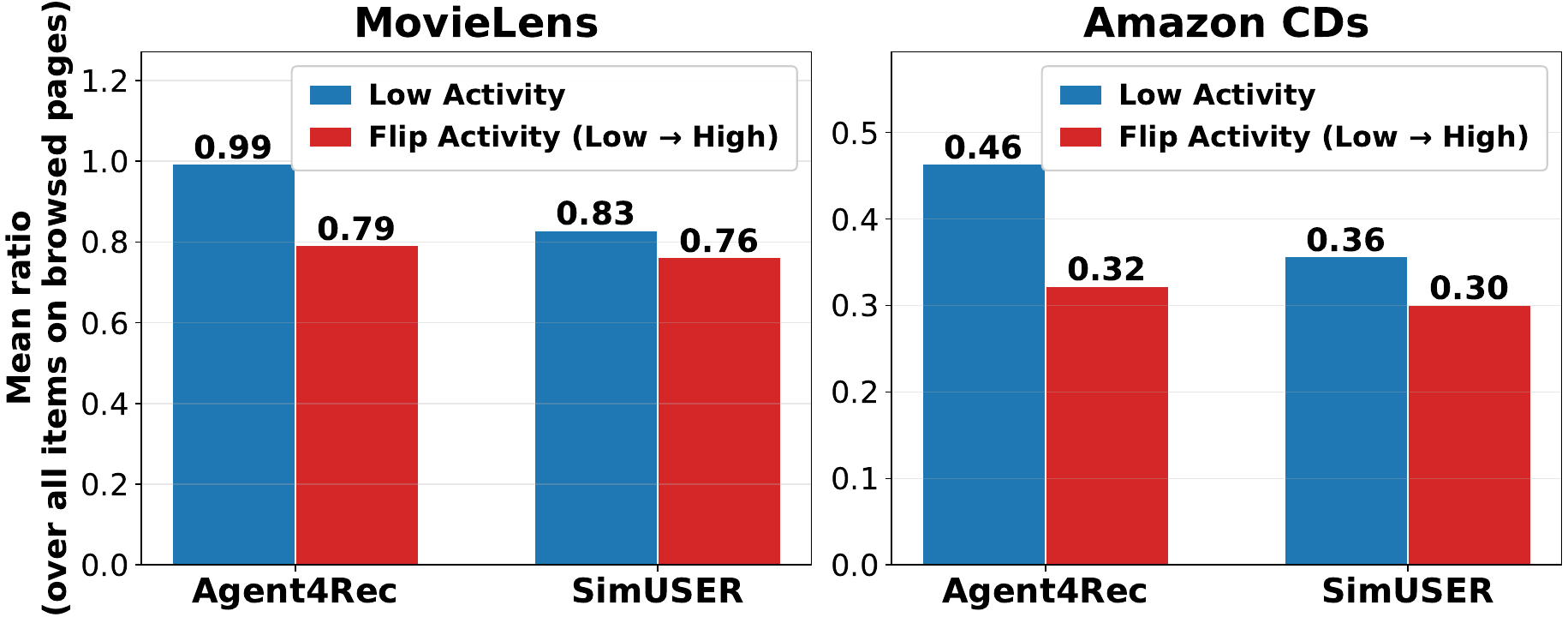}
  \caption{Impact of amplifying the activity trait on the category overlap ratio across MovieLens and CDs.}
    \label{fig: mean_shift}
\end{figure}

Figure~\ref{fig: mean_shift} shows the overlap-to-baseline ratio for each activity condition, where the ratio is computed at the user-page level and averaged over all pages actually interacted with by the simulators.
The low-activity simulator tends to exit prematurely regardless of page quality, restricting its interactions to early, high-matching items and thereby mechanically increasing the mean ratio. However, this high ratio reflects a structural blind spot rather than precise filtering, since the simulator may miss preference-conforming items in later pages due to overly rigid exit behavior.
Conversely, amplifying the activity trait forces the simulator to explore a drastically larger number of pages, causing the mean ratios to drop across both datasets. 
This indicates that the behavioral constraint of extended browsing overrides the simulator's intrinsic preference boundaries, driving it to interact indiscriminately. 
This historical baseline therefore serves not only as a diagnostic metric but also as the basis for mitigating the interference, which we formalize in Section~\ref{sec: methods}.



\subsubsection{\textbf{Correlation between Activity Trait and User Rating}}\label{sec: Correlation between Activity and Rating}
To validate whether the permissive item acceptance under high-activity states reflects natural human behavior, we analyze the correlation between user activity and item ratings using real-world datasets. 
{We find a statistically significant negative correlation between user activity and average historical ratings in both MovieLens (Kendall's $\tau=-0.366$, p-value $=2.9\times10^{-25}$) and CDs (Kendall's $\tau=-0.069$, p-value $=1.65\times10^{-28}$).}
This indicates that high-activity users tend to apply stricter evaluation criteria rather than exhibiting more permissive behavior.
In contrast, amplifying the activity trait makes the simulator relax its preference criteria (Sections~\ref{sec: persona-item embedding similarity},~\ref{sec: category overlap}). This discrepancy indicates that the simulator's permissive behavior is not a natural consequence of high activity, but trait interference induced by the extended-browsing constraint.

\subsection{Evaluation Invalidity}
\label{sec: evaluation invalidity}
The trait interference observed in the previous Section~\ref{sec: trait interference} raises a critical concern regarding the validity of page-level satisfaction metrics. We hypothesize that the satisfaction scores artificially inflate as more pages are explored, regardless of actual recommendation quality. This implies that evaluation outcomes are dominated by the activity trait rather than the recommendation model's actual performance. To verify this, we conduct two analyses: we first compare the activity trait against preference-related traits to verify whether the inflation is specific to activity (Section~\ref{sec: trait comparison}), and then examine whether this inflation persists under systematically controlled recommendation quality (Section~\ref{sec: recommendation quality}). Throughout the experiments, we evaluate the simulation using the average viewing ratio ($\overline{P}_{view}$), the average number of explored pages ($\overline{N}_{exit}$), and the average user satisfaction score ($\overline{S}_{sat}$).

\subsubsection{\textbf{Activity vs. Preference Traits}}
\label{sec: trait comparison}
\begin{table}[]
\centering
\caption{Changes in simulator evaluation metrics when each user trait is amplified from low to high.}
\resizebox{1\linewidth}{!}{
\begin{tabular}{cc|c|ccc|ccc}
\toprule[1.5pt]
\multicolumn{2}{c|}{\multirow{2}{*}{Trait Settings}} & \multirow{2}{*}{Simulator} & \multicolumn{3}{c|}{MovieLens} & \multicolumn{3}{c}{CDs} \\ \cmidrule{4-9} 
\multicolumn{2}{c|}{} & & $\overline{P}_{view}$ & $\overline{N}_{exit}$ & $\overline{S}_{sat}$ & $\overline{P}_{view}$ & $\overline{N}_{exit}$ & $\overline{S}_{sat}$ \\ \midrule\midrule
\multicolumn{1}{c|}{\multirow{4}{*}{Activity}} & \multirow{2}{*}{Low} & Agent4Rec & 0.509 & 1.00 & 3.18 & 0.529 & 1.07 & 3.46 \\
\multicolumn{1}{c|}{} & & SimUSER & 0.409 & 4.44 & 4.85 & 0.369 & 2.73 & 4.85 \\ \cmidrule{2-9} 
\multicolumn{1}{c|}{} & \multirow{2}{*}{Low $\rightarrow$ High} & Agent4Rec & 0.519 & 3.10 & 7.44 & 0.490 & 4.22 & 7.04 \\
\multicolumn{1}{c|}{} & & SimUSER & 0.513 & 4.97 & 7.12 & 0.419 & 4.95 & 6.57 \\ \midrule
\multicolumn{1}{c|}{\multirow{4}{*}{Conformity}} & \multirow{2}{*}{Low} & Agent4Rec & 0.555 & 1.36 & 3.74 & 0.515 & 1.32 & 3.65 \\
\multicolumn{1}{c|}{} & & SimUSER & 0.460 & 2.90 & 4.64 & 0.403 & 3.03 & 5.21 \\ \cmidrule{2-9} 
\multicolumn{1}{c|}{} & \multirow{2}{*}{Low $\rightarrow$ High} & Agent4Rec & 0.547 & 1.32 & 3.86 & 0.585 & 1.42 & 4.15 \\
\multicolumn{1}{c|}{} & & SimUSER & 0.449 & 3.30 & 5.25 & 0.451 & 3.39 & 6.09 \\ \midrule
\multicolumn{1}{c|}{\multirow{4}{*}{Diversity}} & \multirow{2}{*}{Low} & Agent4Rec & 0.510 & 1.10 & 3.34 & 0.507 & 1.13 & 3.47 \\
\multicolumn{1}{c|}{} & & SimUSER & 0.385 & 2.53 & 4.32 & 0.362 & 2.59 & 4.67 \\ \cmidrule{2-9} 
\multicolumn{1}{c|}{} & \multirow{2}{*}{Low $\rightarrow$ High} & Agent4Rec & 0.570 & 1.08 & 3.50 & 0.565 & 1.13 & 3.64 \\
\multicolumn{1}{c|}{} & & SimUSER & 0.465 & 3.29 & 5.05 & 0.439 & 3.26 & 5.59 \\ 
\bottomrule[1.5pt]
\end{tabular}
}
\label{tab: evaluaiton invalidity}
\end{table}

To verify whether this evaluation invalidity is uniquely caused by the activity trait, we compare its impact against preference-related traits (i.e., conformity and diversity), as summarized in Table~\ref{tab: evaluaiton invalidity}. While altering the preference traits yields only marginal fluctuations in $\overline{N}_{exit}$ and $\overline{S}_{sat}$, amplifying the activity trait causes drastic inflations in both metrics. This stark contrast confirms that the evaluation metric is specifically distorted by the behavioral constraint of the activity trait, rather than by preference-related traits.

\subsubsection{\textbf{Inflation under Recommendation Quality}}
\label{sec: recommendation quality}

\begin{table}[]
\centering
\vspace{2ex}
\caption{Impact of amplifying the activity trait on simulation metrics under controlled page qualities. Pages are degraded by varying the positive-to-negative ratio ($1:k$).}
\label{tab: evaluation invalidity 1:k}
\resizebox{1\linewidth}{!}{
\begin{tabular}{cc|c|ccc|ccc}
\toprule[1.5pt]
\multirow{2}{*}{Ratio ($1:k$)} & \multirow{2}{*}{Activity} & \multirow{2}{*}{Simulator} & \multicolumn{3}{c|}{MovieLens} & \multicolumn{3}{c}{CDs} \\ \cmidrule{4-9} 
 & & & $\overline{P}_{view}$ & $\overline{N}_{exit}$ & $\overline{S}_{sat}$ & $\overline{P}_{view}$ & $\overline{N}_{exit}$ & $\overline{S}_{sat}$ \\ \midrule\midrule
\multirow{4}{*}{$1:1$} & \multirow{2}{*}{Low} & Agent4Rec & 0.509 & 1.00 & 3.18 & 0.529 & 1.07 & 3.46 \\
 & & SimUSER & 0.409 & 4.44 & 4.85 & 0.369 & 2.73 & 4.85 \\ \cmidrule{2-9} 
 & \multirow{2}{*}{Low $\rightarrow$ High} & Agent4Rec & 0.519 & 3.10 & 7.44 & 0.490 & 4.22 & 7.04 \\
 & & SimUSER & 0.513 & 4.97 & 7.12 & 0.419 & 4.95 & 6.57 \\ \midrule
\multirow{4}{*}{$1:3$} & \multirow{2}{*}{Low} & Agent4Rec & 0.514 & 1.01 & 3.15 & 0.488 & 1.09 & 3.43 \\
 & & SimUSER & 0.359 & 2.60 & 4.52 & 0.321 & 2.32 & 4.53 \\ \cmidrule{2-9} 
 & \multirow{2}{*}{Low $\rightarrow$ High} & Agent4Rec & 0.462 & 4.14 & 6.99 & 0.434 & 3.91 & 6.63 \\
 & & SimUSER & 0.438 & 4.97 & 6.47 & 0.350 & 4.98 & 6.17 \\ \midrule
\multirow{4}{*}{$1:9$} & \multirow{2}{*}{Low} & Agent4Rec & 0.389 & 1.00 & 3.06 & 0.380 & 1.01 & 3.15 \\
 & & SimUSER & 0.310 & 2.02 & 3.89 & 0.271 & 2.02 & 4.01 \\ \cmidrule{2-9} 
 & \multirow{2}{*}{Low $\rightarrow$ High} & Agent4Rec & 0.415 & 3.89 & 6.51 & 0.396 & 3.77 & 6.34 \\
 & & SimUSER & 0.393 & 4.99 & 6.03 & 0.311 & 5.00 & 5.71 \\ 
\bottomrule[1.5pt]
\end{tabular}
}
\end{table}

To verify that this inflation is independent of recommendation quality, we degrade the pages by increasing $k$ in the $1:k$ ratio (Table~\ref{tab: evaluation invalidity 1:k}), so that more items are sampled from lower predicted logits, and compare metrics as activity is amplified from low to high.
As shown in Table~\ref{tab: evaluation invalidity 1:k}, we have the following observations: 1) Irrespective of the positive-to-negative item ratio, amplifying the activity trait causes significant inflations in both the average number of explored pages ($\overline{N}_{exit}$) and the average satisfaction score ($\overline{S}_{sat}$). 2) Crucially, this artificial inflation persists even at the extreme $1:9$ setting, where negative items explicitly mismatch the user's taste yet the high-activity simulator continues to explore significantly more pages. This confirms that the evaluation metric merely reflects the injected activity trait rather than the true relevance of the recommended items, ultimately rendering the evaluation invalid.

\section{Methods}
\label{sec: methods}

The empirical evidence from Sections~\ref{sec: trait interference} and \ref{sec: evaluation invalidity} reveals a critical flaw in current simulators: activity often determines exit behavior without sufficient regard to page quality. 
When instructed to exhibit high activity, the LLM continues browsing preference-mismatched pages despite degraded recommendation quality; when assigned low activity, it may terminate mechanically even when the page still contains preference-aligned items.
In an ideal simulation, activity should modulate browsing depth only when page quality remains acceptable, rather than overriding preference alignment. 
We attribute this failure to the absence of a clear, personalized standard for determining whether a page remains acceptable to the user. 
To address this issue, we equip the simulator with an explicit page-level quality anchor, ensuring that the activity trait determines exploration depth \emph{only within preference-conforming pages}.

\smallskip
\noindent\textbf{PQA: Page-level Quality Anchor. }
To enforce this ideal behavior, we propose PQA, which injects a personalized page-level quality anchor $\mu_u$ into the LLM's system prompt. Instead of introducing an arbitrary threshold, we formally define $\mu_u$ using the user's historical category overlap $\mathrm{baseline}$ previously formulated in Equation~\ref{eq: baseline} (i.e., $\mu_u := \mathrm{baseline}$).
Because this $\mathrm{baseline}$ is derived directly from the user's real-world interaction history, it provides a natural, data-driven standard that explicitly quantifies their intrinsic preference boundaries.
To operationalize this anchor, we translate the numerical evaluation into an explicit, rule-based labeling system within the LLM's prompt. During the simulation, the category match score of the current page is dynamically compared against $\mu_u$ to calculate a relative quality ratio. 
Based on this ratio, each page is assigned one of three qualitative labels: \emph{ABOVE}, \emph{NORMAL}, or \emph{BELOW}.
\footnote{Dataset-specific ratio thresholds are $(1.0, 0.7)$ for MovieLens and $(0.5, 0.3)$ for CDs; the three intervals define \emph{ABOVE}, \emph{NORMAL}, and \emph{BELOW}, respectively.}
This label serves as a page-level quality signal that gates how the activity trait affects browsing decisions.
The simulator first generates its overall feeling based primarily on page quality, while the activity trait is explicitly prevented from changing this quality assessment. It then makes the continue-or-exit decision by jointly considering the page-quality label, activity trait, and accumulated session fatigue.

{Specifically, for an \emph{ABOVE} page, PQA applies a one-page continuation rule because the page is judged to sufficiently satisfy the user's preference standard. This rule prevents low-activity simulated users from mechanically exiting too early. It does not assume that exiting after finding satisfying items is irrational; rather, it delays only the immediate exit decision by one page. If subsequent pages are labeled \emph{NORMAL} or \emph{BELOW}, activity, fatigue, and repeated low-quality signals are considered again in the exit decision.
A \emph{NORMAL} page triggers a sticky one-page exploration on first encounter to avoid mechanical early exits, and is thereafter judged by activity and fatigue. A \emph{BELOW} page is granted a single grace page only when it appears on the first page, to reflect minimal curiosity. Two consecutive \emph{BELOW} pages force exit regardless of activity; in all other cases, low activity increases exit sensitivity while high activity allows limited patience. Thus, activity modulates browsing depth only after page quality has been explicitly assessed, preventing it from overriding the user's core preferences.}


\section{Experiments}

\begin{figure}[t]
  \centering
  \includegraphics[width=1\linewidth]{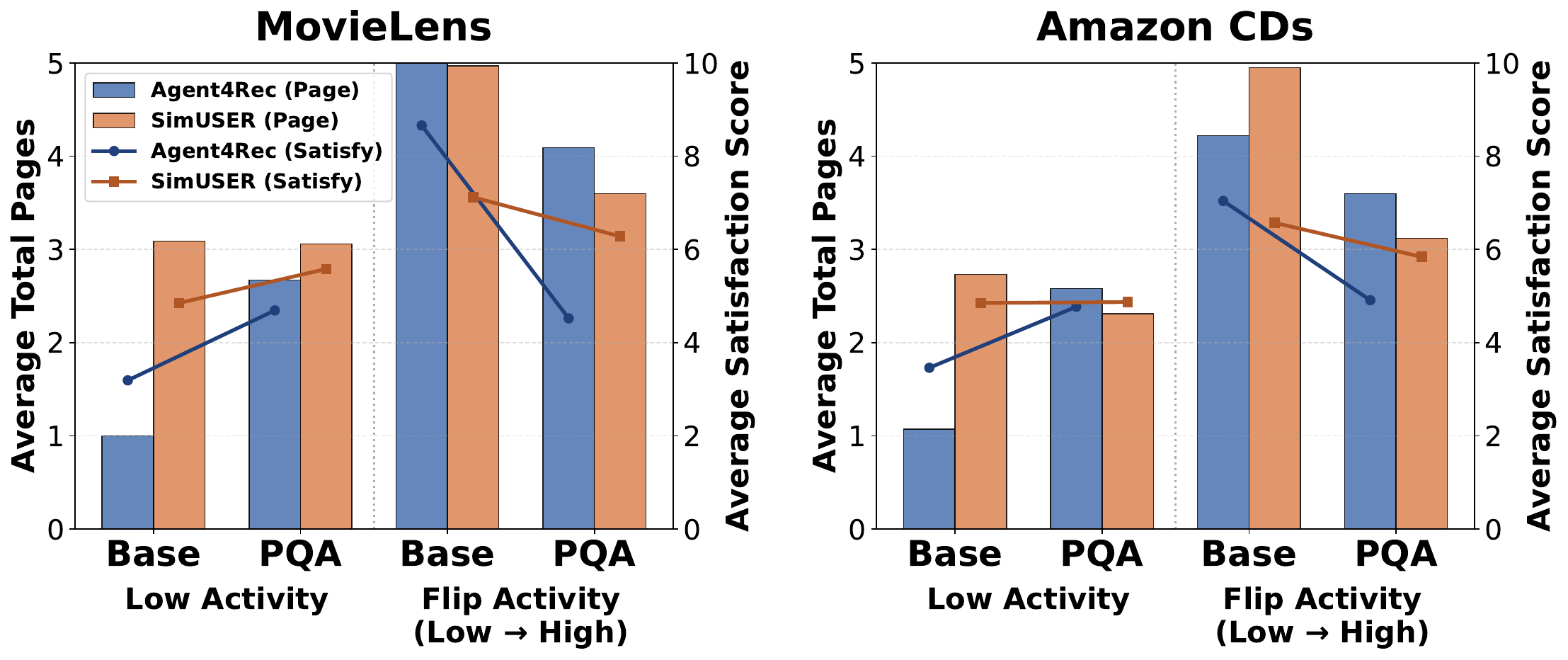}
  \caption{Effect of PQA on browsing depth and satisfaction scores under activity shifts on MovieLens and CDs.}
    \label{fig: page_satisfy}
\end{figure}

\begin{table}[]
\vspace{2ex}
\centering
\caption{Recommendation results after fine-tuning with simulator-generated interactions on MovieLens and CDs.}
\label{tab:results}
\resizebox{1\linewidth}{!}{
\begin{tabular}{cc|c|cc|cc}
\toprule[1.5pt]
\multirow{2}{*}{Variant} & \multirow{2}{*}{Trait Settings} & \multirow{2}{*}{Simulator} & \multicolumn{2}{c|}{MovieLens} & \multicolumn{2}{c}{CDs} \\ \cmidrule{4-7}
 & & & NDCG@5 & HR@5 & NDCG@5 & HR@5 \\ \midrule\midrule
\multirow{4}{*}{Base} & \multirow{2}{*}{Low} & Agent4Rec & 0.2819 & 0.3915 & 0.3322 & 0.4623 \\
 & & SimUSER & 0.2945 & 0.4026 & 0.3187 & 0.4390 \\ \cmidrule{2-7}
 & \multirow{2}{*}{Low $\rightarrow$ High} & Agent4Rec & 0.2945 & 0.4007 & 0.3273 & 0.4519 \\
 & & SimUSER & 0.2842 & 0.3952 & 0.3200 & 0.4390 \\ \midrule
\multirow{4}{*}{PQA} & \multirow{2}{*}{Low} & Agent4Rec & \textbf{0.3046} & \underline{0.4082} & \textbf{0.3479} & \textbf{0.4805} \\
 & & SimUSER & 0.3004 & 0.3933 & \underline{0.3437} & \underline{0.4675} \\ \cmidrule{2-7}
 & \multirow{2}{*}{Low $\rightarrow$ High} & Agent4Rec & 0.2950 & 0.4063 & 0.3341 & 0.4623 \\
 & & SimUSER & \underline{0.3024} & \textbf{0.4156} & 0.3264 & 0.4519 \\
\bottomrule[1.5pt]
\end{tabular}
}
\end{table}


\begin{table}[t]
\vspace{2ex}
\centering
\scriptsize 
\caption{Human-likeness score evaluated by GPT-4o across recommendation domains.}
\label{tab:human_likeness}
\resizebox{0.7\linewidth}{!}{%
\begin{tabular}{lcc}
\toprule
 & {MovieLens} & {CDs} \\ 
\midrule
Agent4Rec & $3.57 \pm 0.03$ & $3.64 \pm 0.02$ \\
SimUSER & $4.26 \pm 0.04$ & $4.05 \pm 0.11$ \\ 
\midrule
PQA-Agent4Rec & $4.15 \pm 0.04$ & $\mathbf{4.22 \pm 0.04}$ \\
PQA-SimUSER & $\mathbf{4.35 \pm 0.08}$ & $4.06 \pm 0.07$ \\
\bottomrule
\end{tabular}%
}
\end{table}

We conduct experiments on two widely used datasets: MovieLens~\cite{harper2015movielens} and Amazon CDs~\cite{mcauley2015image}. 
To evaluate the reliability of trait-driven behaviors, we apply our framework to two LLM-based user simulators, Agent4Rec~\cite{zhang2024generative} and SimUSER~\cite{bougie2025simuser}. Unless otherwise stated, all simulations are conducted using GPT-4o-mini as the underlying LLM.

\smallskip
\noindent\textbf{1. Restoring Trait Independence and Evaluation Validity under Activity Shifts. }
To examine whether PQA restores the independence between preference and behavioral traits, {we use a controlled 1:1 recommended list consisting of 20 items, partitioned into five pages with four items per page. Since the first half of the list is preference-aligned and the second half is preference-mismatched, Page 3 forms the preference boundary, while Pages 4--5 contain only low-quality items.} 
As shown in Figure~\ref{fig: page_satisfy}, baseline simulators exhibit rigid browsing patterns driven by the injected activity level. Across both datasets, Agent4Rec often exits prematurely under low activity despite remaining preference-aligned items, whereas both baselines over-browse low-quality pages under high activity.
The corresponding increase in satisfaction scores under high activity indicates evaluation invalidity, where satisfaction reflects the activity level more than actual recommendation quality.
PQA mitigates these failures by grounding browsing decisions in the personalized quality anchor $\mu_u$. 
Under low activity, it reduces premature exits by identifying pages that remain preference-aligned, leading Agent4Rec to explore more relevant items and report higher satisfaction. 
Under high activity, it prevents activity from becoming an unconditional mandate to continue browsing, reducing excessive exploration and satisfaction inflation once page quality falls below the user's preference standard.
SimUSER browses more deeply under low activity due to its exit-refinement procedure, yet still over-browses under high activity; PQA mitigates this by grounding exits in page-level alignment, reducing trait interference and improving evaluation reliability under activity shifts.


\smallskip
\noindent\textbf{2. Downstream Validation via Simulated Interaction Augmentation. }
To assess the fidelity of PQA-generated interactions, we use the simulation logs for data augmentation and evaluate whether they improve downstream recommendation performance.
Specifically, we collect virtual interactions from Agent4Rec and SimUSER, with or without PQA, and use them as new positive preference labels to fine-tune SASRec~\cite{kang2018self}.
We then compare the fine-tuned models on the offline test set, under the hypothesis that a simulator aligned with users' authentic preferences yields a stronger augmented model on real test data.
As shown in Table ~\ref{tab:results}, models fine-tuned with PQA-generated interactions achieve the best performance across both datasets.
In particular, PQA-Agent4Rec obtains the largest gains, reaching an \text{NDCG@5} of 0.3046 on MovieLens and 0.3479 on CDs.
These results indicate that PQA effectively filters low-quality pages and preserves the user's intrinsic preferences. While high-activity baseline simulators keep consuming preference-mismatched items in later pages and inject them into the simulated data, PQA uses $\mu_u$ to terminate browsing before such interactions accumulate, yielding cleaner data that better reflects users' genuine preferences.

\smallskip
\noindent\textbf{3. LLM-based Human-Likeness Evaluation. }
We use GPT-4o to evaluate the human-likeness of simulator-generated interactions on a 5-point Likert scale~\cite{chiang2023can}, with higher scores indicating closer alignment with real user behavior.
Table~\ref{tab:human_likeness} shows that PQA consistently improves both simulators across datasets. PQA-Agent4Rec increases the score from $3.57$ to $4.15$ on MovieLens and from $3.64$ to $4.22$ on CDs, while PQA-SimUSER achieves the highest score of $4.35$ on MovieLens.
The improvement stems from PQA's context-aware exit mechanism. Without PQA, simulators often exhibit rigid, suboptimal browsing patterns, either exiting too early under low activity or continuing through low-quality pages under high activity, which lowers human-likeness. 
In contrast, by grounding exit decisions in $\mu_u$, PQA enables simulators to exit on taste-mismatched pages while continuing when recommendations remain acceptable, producing more realistic and preference-faithful interaction logs.

\section{Related Work}
LLM-based user simulation has emerged as a promising alternative to static
recommendation metrics and costly online A/B testing. 
Offline metrics evaluate recommendation systems on fixed logged interactions and thus cannot capture how users respond to newly exposed items, while online A/B testing is costly and risks
degrading user experience.
Early work in this direction has primarily focused on constructing realistic user personas.
Agent4Rec~\cite{zhang2024generative} models social traits such as conformity
from real-world data, while profile-aware simulators~\cite{fabbri2025evaluating}
summarize user history in natural language to better align with human judgment.
PUB~\cite{ma2025pub} further integrates Big Five personality
traits~\cite{Goldberg1992THEDO, Roccas2002TheBF} to replicate diverse
interaction patterns, and SimUSER~\cite{bougie2025simuser} identifies
self-consistent personas with specialized perception and memory modules to serve
as believable human proxies.
Furthermore, AgentCF~\cite{zhang2024agentcf} places the collaborative signal inside the
simulator, casting both users and items as agents and jointly optimizing their
textual memories to fit observed user-item interactions.
Recent work shifts focus from constructing simulators to using them as a source
of feedback for the recommender. RecoWorld~\cite{liu2026recoworld} establishes a
proactive feedback loop in which the simulator explicitly signals user states
(e.g., boredom) to guide the recommender's adaptation.
Despite this progress, whether simulators behave as specified has received little attention, as evaluation typically reports aggregate satisfaction that conceals whether the injected traits retained their intended roles.
Existing simulators implicitly assume trait independence, where preference attributes determine what a user engages with while a behavioral activity trait governs only how long the user browses. We show that this assumption collapses during simulation, and that the resulting interference propagates into satisfaction-based evaluation.

\section{Conclusion}
{In this paper, we identified trait interference as a failure mode of LLM-based user simulators, where activity traits interfere with preference judgments and undermine satisfaction-based evaluation. We proposed PQA, which grounds exit decisions in personalized page-level preference alignment. Experiments show that PQA reduces activity-driven over-browsing, restores trait independence, and improves the reliability of simulator-generated interactions.
A promising direction is to extend PQA beyond category-level signals with richer preference indicators (e.g., item-item relationships, collaborative filtering signals, or fine-grained feedback) to better capture users' intrinsic preferences across domains.

\begin{acks}
This work was supported by the National Research Foundation of Korea (NRF) grant funded by the Korea government (MSIT) (RS-2024-00335098), the National Research Foundation of Korea (NRF) grant funded by the Korea government (MSIT) (RS-2024-00406985), and partly supported by Institute for Information \& Communications Technology Planning \& Evaluation (IITP) grant funded by the Korea government (MSIT) (RS-2019-II190075), Artificial Intelligence Graduate School Support Program (KAIST).
\end{acks}


\clearpage

\section*{GenAI Usage Disclosure}
We acknowledge the use of generative AI tools, such as Gemini and Claude, exclusively for limited assistance in checking grammar, improving readability, refining expression, and reducing length to satisfy page constraints. We also used these tools for minor refactoring and debugging of code related to plotting and visualization. All AI-assisted revisions and code modifications were carefully reviewed and validated by the authors. The core ideas, methodology, experiments, and interpretations presented in this paper are entirely the authors' original contributions.

\bibliographystyle{ACM-Reference-Format}
\balance
\bibliography{sample-base}

\appendix

\end{document}